\documentclass[letterpaper, 10 pt, conference]{ieeeconf} 
\IEEEoverridecommandlockouts
\usepackage{graphics} 
\usepackage{epsfig} 
\usepackage{amsmath} 
\usepackage{amssymb} 
\usepackage{color}
\usepackage{bm}
\usepackage{bbm}
\usepackage{subcaption}
\usepackage{multirow}
\usepackage{array}
\usepackage{booktabs}
\usepackage{algorithm,algcompatible}
\usepackage{mathtools}
\usepackage{cite}
\usepackage{hyperref}
\usepackage{cuted}
\definecolor{pointcolor}{RGB}{0,114,189}
\definecolor{linecolor}{RGB}{217,83,25}
\definecolor{planecolor}{RGB}{237,177,32}
\definecolor{spherecolor}{RGB}{119,172,48}
\definecolor{ellipsoidcolor}{RGB}{126,47,142}
\definecolor{cylindercolor}{RGB}{77,190,238}
\definecolor{conecolor}{RGB}{162,20,47}

\graphicspath{{figures/}}

\title{\huge \bf LiDAR-enhanced 3D Gaussian Splatting Mapping}

\author{Jian Shen\quad  Huai Yu$^\dagger$ \quad  Ji Wu\quad Wen Yang\quad  Gui-Song Xia$^\dagger$
\thanks{$^\dagger$Corresponding authors.}
\thanks{Jian Shen and Ji Wu are with the School of Computer Science, Huai Yu and Wen Yang are with the School of Electronic Information, Gui-Song Xia is with the School of Artificial Intelligence, National Engineering Research Center for Multimedia Software, and Institute for Math \& AI, Wuhan. All authors are with Wuhan University, Wuhan, China 430072. E-mail: {\tt\small\{jianshen, yuhuai, ji.wu, yangwen, guisong.xia\}@whu.edu.cn}}
}

\begin{document}

\maketitle
\begin{abstract}
This paper introduces LiGSM, a novel LiDAR-enhanced 3D Gaussian Splatting (3DGS) mapping framework that improves the accuracy and robustness of 3D scene mapping by integrating LiDAR data. LiGSM constructs joint loss from images and LiDAR point clouds to estimate the poses and optimize their extrinsic parameters, enabling dynamic adaptation to variations in sensor alignment. Furthermore, it leverages LiDAR point clouds to initialize 3DGS, providing a denser and more reliable starting points compared to sparse SfM points. In scene rendering, the framework augments standard image-based supervision with depth maps generated from LiDAR projections, ensuring an accurate scene representation in both geometry and photometry. Experiments on public and self-collected datasets demonstrate that LiGSM outperforms comparative methods in pose tracking and scene rendering.


\end{abstract}
\section{Introduction}

3D Gaussian Splatting (3DGS) is a technology used for 3D scene modeling and rendering, achieving an efficient rendering process by representing the scene as a set of learnable 3D Gaussian functions \cite{kerbl20233d}. It has applications in various fields such as 3D mapping \cite{keetha2024splatam} and autonomous driving \cite{zhou2024drivinggaussian}. However, pure 3DGS relying solely on visual information struggles to accurately initialize the 3D space due to its dependence on sparse and potentially unreliable point clouds from Structure from Motion (SfM).


LiDAR provides precise 3D information, unaffected by the lighting conditions and capable of capturing intricate environmental details over a much longer effective range compared to depth cameras, making it ideal for enhancing the geometric fidelity of reconstructed scenes. By incorporating LiDAR measurements, we can achieve more precise initialization of the 3DGS, thereby improving the overall quality and reliability of the reconstructed geometry. This fusion strategy also mitigates the dependency on sparse initialization typical in purely visual approaches, thus potentially increasing the robustness of the reconstruction process.

\begin{figure}[ht]
    \centering
    \includegraphics[width=1\linewidth]{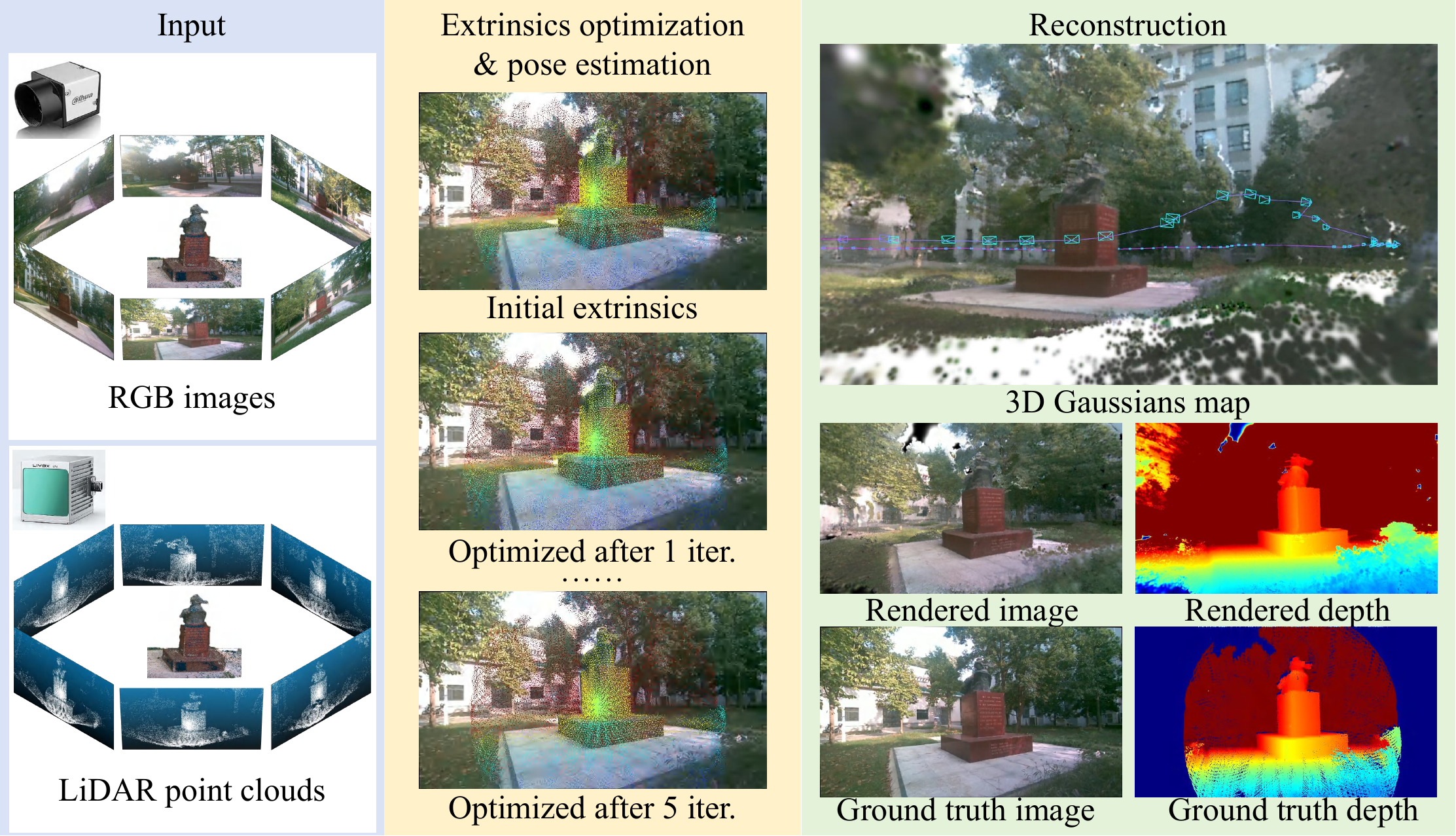}
    \caption{LiGSM takes images and LiDAR point clouds as inputs, simultaneously optimizing the LiDAR-camera extrinsics and estimating the poses. The 3DGS reconstruction results is capable of rendering color images and depth maps.}
    \label{fig:intro}
\end{figure}

Integrating LiDAR into visual 3DGS brings advantages but also new challenges. The first challenge lies in how to effectively fuse LiDAR data within the 3DGS rendering pipeline. Traditional 3DGS methods rely heavily on image-based cues for both training and inference, but integrating LiDAR requires developing novel mechanisms to incorporate depth information seamlessly alongside color data. This necessitates modifications to the existing architecture and loss functions to ensure that the LiDAR data contributes meaningfully to the optimization process without overshadowing the visual cues that are essential for texturing and appearance synthesis. The second challenge involves ensuring and continuously maintaining accurate extrinsic calibration between the LiDAR and the camera. In real-world scenarios, even minor misalignments or variations in the relative positioning of sensors can lead to significant errors in the final mapping. Moreover, environmental changes or modifications in sensor mounting can cause alignment shifts over time, necessitating dynamic recalibration methods. Addressing this issue requires a robust system capable of detecting and correcting any drift or misalignment during operation.

In this work, we propose a novel mapping framework that leverages LiDAR-enhanced 3DGS, designed to improve the accuracy and robustness of 3D scene mapping, as shown in Fig.~\ref{fig:intro}. We optimize the extrinsic parameters by constructing a joint observation of LiDAR point clouds and images, while simultaneously optimizing the pose. This joint optimization allows our system to adapt dynamically to variations in sensor alignment, ensuring that the reconstruction remains accurate even when the relative positions of the camera and LiDAR change. Furthermore, we leverage LiDAR point clouds for initializing the 3DGS, providing a denser and more reliable starting point than traditional sparse feature matches. For scene rendering, we augment the standard image-based supervision with depth maps generated from LiDAR projections. This dual-supervision strategy ensures that both the geometric and photometric aspects of the scene are accurately represented, leading to higher-quality reconstructions. The key contributions are summarized as:


\begin{itemize}
\item We propose LiGSM, which leverages LiDAR to enhance the performance of visual 3D scene mapping.
\item We design a novel pose estimation framework fusing LiDAR and a monocular camera, which can optimize the camera-LiDAR extrinsic parameter without re-calibration when estimating poses.

\item We integrate the geometric information of LiDAR data into the initialization and training process of 3DGS, which reduces the dependence on the depth of visual reconstruction and improves the robustness of the reconstruction on image quality. 
\item Experiments conducted on both public and self-collected datasets demonstrate the robustness and effectiveness of the proposed LiGSM on extrinsic-updating and dense mapping.
\end{itemize}

\section{Related Work}

\subsection{Dense Scene Representation for SLAM}

Dense 3D scene representation is typically based on grids, points and neural networks. Grid-based methods \cite{choi2015robust, li2022bnv, sun2021neuralrecon, weder2020routedfusion, weder2021neuralfusion, zhou2013dense} manage memory effectively by using dense grids, hierarchical octrees, and voxel hashing. Although they are simple and fast for finding neighborhoods and integrating context, they require predefined resolution, which can lead to inefficient memory usage in sparse areas and fail to capture details. Point-based methods \cite{cao2018real, cho2021sp, chung2023orbeez, kahler2015very, keller2013real, schops2019bad, zhang2020dense} address the limitations of grids by varying density, focusing points on surfaces, avoiding modeling of empty space, but struggle to model continuous scenes. Neural methods \cite{azinovic2022neural, mescheder2019occupancy, sucar2021imap, wang2023co, wang2022neuris, yan2021continual} implicitly model continuous scenes through coordinate networks, capable of capturing high-quality maps and textures, but network training is time-consuming and not suitable for online reconstruction. Recently, new representation methods such as surfels \cite{gao2023surfelnerf, mihajlovic2021deepsurfels} and neural planes \cite{johari2023eslam, peng2020convolutional} have been proposed, but they have limitations in modeling flexibility. NeRF \cite{mildenhall2021nerf} uses neural radiance fields to represent scenes, effectively modeling continuous scenes, and can render realistic images from neural radiance fields through volumetric rendering. However, training and rendering of neural radiance fields involve inevitable sampling of empty areas, which is computationally expensive, especially when dealing with large-scale scenes. Kerbl \textit{et al.} proposed the use of 3DGS \cite{kerbl20233d}, representing volumetric elements in the scene with three-dimensional Gaussian functions. Compared to NeRF, 3DGS has a significant advantage in rendering and training speed because it does not need to represent empty areas, and it allows for more direct scene editing and manipulation. However, pure visual 3DGS rely on the accuracy of sparse point clouds generated by SfM for initialization, which can affect the quality of the final 3D reconstruction.

\subsection{Multimodal Dense 3D Reconstruction}

Recently, numerous SLAM studies have adopted neural radiance fields and 3D Gaussians as methods for 3D scene representation \cite{sucar2021imap, rosinol2023nerf, wang2023co, zhu2022nice, zhu2024nicer, huang2024photo, hong2024liv, lang2024gaussian, keetha2024splatam}. Initially, these works utilized images as inputs \cite{zhu2024nicer}, employing gradient descent algorithms to optimize both the scene and poses through differentiable rendering. However, supervision using images alone often yields subpar results. SLAM algorithms that use RGB-D inputs\cite{wang2023co, zhu2022nice}, which incorporate depth supervision for rendering depth maps, can better leverage geometric information, leading to reconstructions that boast both superior visual appearance and robust structure. Yet, pose calculation based on gradient descent is highly sensitive to the accuracy of initial values, prompting the development of algorithms \cite{huang2024photo} that integrate traditional SLAM front-ends. Subsequently, SLAM algorithms that fuse LiDAR, inertial, and camera data have emerged. By exploiting the complementary characteristics, LIV-GaussMap \cite{hong2024liv} and Gaussian-LIC \cite{lang2024gaussian} can capture the geometric structure of large-scale 3D scenes and recover their visual surface information with high fidelity. However, during the optimization of the 3DGS scene parameters, the geometric information provided by the LiDAR was not utilized for supervision.

\section{Methodology}

\begin{figure}
    \centering
    \includegraphics[width=1\linewidth]{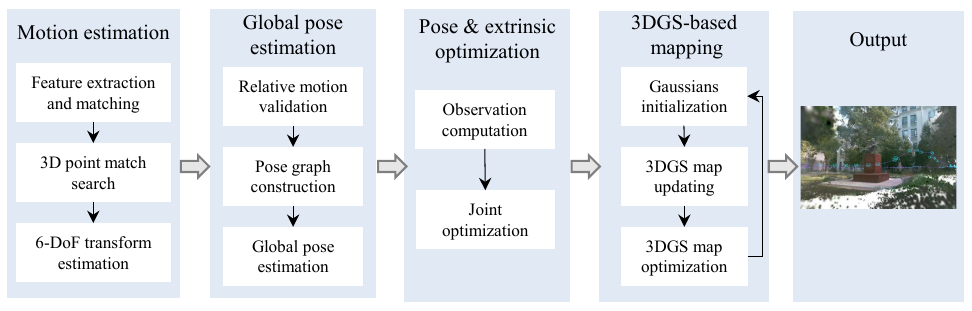}
    \caption{The pipeline of LiGSM. Point clouds are first used for poses and LiDAR-camera extrinsic estimation, and then to initialize 3DGS and optimize the 3D scene mapping.}
    \label{fig:pipeline}
\end{figure}

The introduced LiGSM system utilizes images and corresponding LiDAR point clouds to produce 3D environmental models formatted in 3DGS.
Fig. \ref{fig:pipeline} shows the overview of our LiGSM, described in this section.

\subsection{Relative Motion Estimation}
Given the images, computing the correspondences includes feature extraction, matching, and geometric
verification. Using the extrinsic parameters between the LiDAR and camera, we project the point cloud onto the image plane and search for the 3D points corresponding to the 2D feature points based on the nearest neighbor relationship. Then we use RANSAC+SVD algorithm to estimate the relative motion between two sets of matched 3D points.

\subsection{Global Poses Estimation}
Based on the relative motions, a pose graph is constructed to compute the global poses, where nodes represent poses and edges represent relative motions. 

Before computing the global poses, we validate the relative motions by checking the cycle consistency of the relative motions. 
Specifically, a cycle composed of frames $i$, $j$, and $k$, with their relative motions being $\mathbf{T}_{ij}$, $\mathbf{T}_{jk}$ and
$\mathbf{T}_{ki}$, respectively.  We compute the transformation matrix $\mathbf{T}_{ki} \mathbf{T}_{jk} \mathbf{T}_{ij}$ and evaluate the angle and translation. 
Threshold is used to determine whether a check passes or fails. 
We compute the success rate $R_{success}$ of each relative motion as
\begin{equation}
    R_{success} = \frac{N_{passed}}{N_{involved}},
\end{equation}
where $N_{passed}$ represents the number of passed checks and $N_{involved}$ represents the number of involved checks. Relative motions with a success rate $R_{success}$ below the threshold are considered invalid.

With the valid relative motions $\mathcal{E} = \{\mathbf{T}_{ij}\}$ provided, the initialization of global poses $\mathcal{T} = \{\mathbf{T}_i\}$ can be straightforwardly accomplished by solving the pose graph, where the relative motions act as constraints. We optimize the global poses $\mathcal{T}$ by considering all relative motions in $\mathcal{E}$ and minimizing the following cost function:
\begin{equation}
    \arg\min_{\mathcal{T}} \sum_{\mathbf{T}_{ij} \in \mathcal{E}} \| \text{log}(\mathbf{T}_{ij} \mathbf{T}_j \mathbf{T}_i^{-1}) \|^2,
\end{equation}
where the function $\text{log}$ maps a rigid body transformation matrix to a six-dimensional twist vector in the Lie algebra.

\subsection{Poses and Extrinsics Optimization}
Vision-based SfM algorithms typically use Bundle Adjustment (BA) for poses optimization. Poses estimation is susceptible to long-distance drift, particularly in the absence of effective loop closures. This drift is often exacerbated by factors such as calibration inaccuracies, scale uncertainty, matching errors, and imprecise feature point locations. 
To address these problems, we use LiDAR's geometric information to constrain camera motion and perform joint optimization of the losses from both the camera and LiDAR.

\subsubsection{Camera Loss}
The camera loss is defined as the difference between the projection of 3D feature points onto the image plane and the 2D feature points:

\begin{equation}
    \mathcal{L}_c = \sum_{\mathbf{x}} ||\mathbf{P}\mathbf{T}_i\mathbf{x} - \mathbf{u}||^2,
\end{equation}
where $\mathbf{x} \in \mathbb{R}^{4}$ represents the 3D feature point in homogeneous coordinates, which is obtained through triangulation, $\mathbf{P} \in \mathbb{R}^{3 \times 4}$ is the camera projection matrix, and $\mathbf{u} \in \mathbb{R}^3$ represents the homogeneous representation of the image feature point. 

\subsubsection{LiDAR Loss}
The LiDAR loss is defined based on matched pairs. Each pair consisting of a key point $\mathbf{p} \in \mathbb{R}^4$ in the $i$-th point cloud and a local plane $(\mathbf{q}, \mathbf{n}) \in (\mathbb{R}^4, \mathbb{S}^2)$ in the $j$-th point cloud. $\mathbf{q}$ is the nearest neighbor of $\mathbf{p}$ in the $j$-th point cloud, and $\mathbf{n}$ is the normal vector of the local plane. Both $\mathbf{p}$ and $\mathbf{q}$ are represented in homogeneous coordinates. The LiDAR loss $\mathcal{L}_l$ is defined as

\begin{equation}
    \mathcal{L}_l = \sum_{\mathbf{p}}([\mathbf{n}^T ,0]\cdot(\mathbf{T}_\mathbf{e}^{-1}\mathbf{T}_j\mathbf{T}_i^{-1}\mathbf{T}_e \mathbf{p} - \mathbf{q})),
\end{equation}
where $\mathbf{T}_e$ represents the extrinsic transformation of the LiDAR with respect to the camera. $\mathbf{T}_e$ is adjusted during the optimization process alongside the poses.

\subsubsection{Joint Loss}
Similar to the LiDAR loss, the joint loss is defined based on point-plane matches, except that the points here come from  the triangulation of image features, rather than a LiDAR point cloud. 
For each point $\mathbf{x}$, we examine its multiple views to identify point clouds that capture the same structure. Subsequently, for each identified point cloud, $\mathbf{x}$ is paired with the nearest local plane $(\mathbf{q}, \mathbf{n})$, where $\mathbf{q}$ denotes the point in the point cloud closest to $\mathbf{x}$, and $\mathbf{n}$ represents the normal vector of the local plane. The joint loss $\mathcal{L}_j$ is defined as

\begin{equation}
    \mathcal{L}_j = \sum_{\mathbf{x}}([\mathbf{n}^T ,0] \cdot(\mathbf{T}_e^{-1}\mathbf{T}_i \mathbf{x} - \mathbf{q})).
\end{equation}

\subsubsection{Joint Optimization}
To achieve the joint optimization of poses and extrinsics, we designed a joint cost function composed of previous losses:

\begin{equation}
    \arg\min_{\mathbf{T}_e, \mathcal{T}}
    \lambda_c\mathcal{L}_c+
    \lambda_l\mathcal{L}_l+
    \lambda_j\mathcal{L}_j,
\end{equation}
where $\lambda_c$, $\lambda_l$, and $\lambda_j$ are the weighting factors.

By minimizing the cost function to adjust the camera poses and extrinsics $\mathbf{T}_e$. Since the calculation of relative poses depends on the extrinsics $\mathbf{T}_e$, multiple iterations are required to optimize the extrinsics. In our experiments, the extrinsics $\mathbf{T}_e$ usually converge after 5 to 10 iterations.

\subsection{3DGS-based Mapping}
To achieve realistic image rendering, we use a collection of 3D Gaussians to represent the scene. In comparison to the original 3DGS, we utilize view-independent colors and isotropic Gaussian to reduce parameters. Each Gaussian's parameters include position $\bm{\mu}\in\mathbb{R}^3$, color $\mathbf{c}\in\mathbb{R}^3$, opacity $o\in [0,1]$, and radius $r$.

Given the camera pose $\mathbf{T}$ and the projection matrix $\mathbf{P}$, we can project the 3D Gaussians onto the image plane to obtain the 2D Gaussians. The parameters of the 2D Gaussians are calculated by the following equation:

\begin{equation}
    \left[
	\begin{matrix}
		\bm{\mu}' \\
		1
	\end{matrix}
\right] = \mathbf{P}\mathbf{T} 
\left[
	\begin{matrix}
		\bm{\mu} \\
		1
	\end{matrix}
\right], 
    r' = \frac{fr}{d},
\end{equation}
where $f$ is the focal length, and $d$ is the 3D Gaussian's depth to image plane. The influence weight of the projected 2D Gaussian on pixel point $\mathbf{y}=[u, v]^T$ is given by the following equation:

\begin{equation}
    \alpha = o\exp{(-\frac{||\bm{\mu'}-\mathbf{y}||^2}{2r'})}.
\end{equation}

During rendering, we sort the Gaussians based on their depth to the image plane in ascending order, and perform alpha blending on the colors:

\begin{equation}
\label{eqn:render}
    C(\mathbf{y}) = \sum_{i=1}^n\mathbf{c}_i\alpha_i\Pi_{j=1}^{i-1}(1-\alpha_j).
\end{equation}

Similarly, silhouette images and depth maps can be rendered, as in SplaTAM \cite{keetha2024splatam}
\begin{equation}
    D(\mathbf{y}) = \sum_{i=1}^nd_i\alpha_i\Pi_{j=1}^{i-1}(1-\alpha_j),
\end{equation}

\begin{equation}
    S(\mathbf{y}) = \sum_{i=1}^n\alpha_i\Pi_{j=1}^{i-1}(1-\alpha_j).
\end{equation}

Given estimated camera poses, images and LiDAR point clouds as inputs, we incrementally build the map, performing initialization, updating the map, and global optimization sequentially for each input frame.

\subsubsection{Initialization}
Initialize Gaussians for the current frame using the LiDAR point cluods and color image from the input frame. Specifically, to initialize 3D Gaussians, LiDAR points are projected onto the image plane. Points that project within the image plane are retained. The color of the 3D Gaussian is set to the color of the corresponding projection point on the image. The position of the 3D Gaussian is set to the coordinates from the point cloud. The opacity is set to 0.5, just as in SplaTAM \cite{keetha2024splatam}, and the radius is determined as $r=d/f$.

\subsubsection{Map Updating}
Map updating involves adding areas to the map that are present in the current frame but not yet represented within it. By rendering silhouette images, we can quantify the extent to which the 3D Gaussian map influences each pixel of the current frame. For pixels where this influence falls below a threshold of 0.99, it is inferred that the map lacks a corresponding representation for these areas. Consequently, 3D Gaussians are added in such unrepresented regions of the map.

\subsubsection{Map Optimization}
After updating the map, we proceed with optimizing the parameters within it to construct a globally consistent map. Instead of solely focusing on the current frame, we select a keyframe every five frames for participation in global map optimization. The frames involved in the optimization are shuffled to ensure an unbiased processing order. Optimization of the Gaussian scene parameters is achieved by minimizing the loss $\mathcal{L}_{maping}:$ 

\begin{equation}
    \mathcal{L}_{maping} = \lambda_{color}\mathcal{L}_{color}+
    \lambda_{depth}\mathcal{L}_{depth}.
\end{equation}

We use the color images $\mathbf{I}$ from the input frames to supervise the rendered images $\hat{\mathbf{I}}$, which are rendered according to equation (\ref{eqn:render}). The color loss $\mathcal{L}_{color}$ is formulated as:
\begin{equation}
    \mathcal{L}_{color} = (1-\lambda_{ssim})\cdot ||\hat{\mathbf{I}}-\mathbf{I}||_1+
    \lambda_{ssim} (1-\text{SSIM}(\hat{\mathbf{I}}, \mathbf{I})),
\end{equation}
where \text{SSIM} is used to calculate the structural similarity of images\cite{wang2004image}.

Additionally, we project the point clouds from the input frames onto the image plane to obtain depth maps $\mathbf{D}$, which we then use to supervise the rendered depth maps $\hat{\mathbf{D}}$. Due to the sparsity of the LiDAR depth projection, we use a depth mask $\mathbf{M}_{depth}$ to indicate whether a pixel has a vaild depth value. The depth loss is then computed as:
\begin{equation}
    \mathcal{L}_{depth} =  ||\mathbf{M}_{depth} \cdot(\mathbf{\hat{D}}-\mathbf{D})||_1.
\end{equation}

This dual supervision approach ensures that both the color information and depth perception in the rendered outputs are accurate and consistent with real-world observations.

\section{Experiments}
\label{sec:experiments}
In this section, we detail the experimental datasets,  method configurations, evaluation metrics, and comparisons to the state-of-the-art. Furthermore, we build a sensor pod to collect data for generalization demonstration. 


\subsection{Datasets}
Currently, there is no public dataset for 3D scene mapping and rendering with images and point clouds. Recent methods for this task are often conducted on RGB-D images. To ensure a fair comparison, we also use the RGB-D image dataset. Although LiGSM is designed for paired point clouds and RGB images, it is also applicable to transfer depth images to point clouds using camera intrinsic parameters. 
We first select the public TUM RGB-D dataset \cite{sturm2012tum} and Replica dataset \cite{straub2019replica} for evaluation. The TUM RGB-D dataset \cite{sturm2012tum} contains 39 sequences in an office environment and an industrial hall with RGB images, depth images, and ground truth trajectories. The dataset covers various camera motions that suffer from poor quality with motion blur, on which we test the tracking performance and rendering quality. The synthetic dataset Replica \cite{straub2019replica} includes a variety of indoor scenes, encompassing high-quality RGB images and precise depth maps, on which we evaluate the rendering quality. To improve the running efficiency and demonstrate the effectiveness of our method on 3D mapping, we perform sampling on the public datasets. For the TUM RGB-D dataset, we select scenes fr1/desk, fr1/desk2, fr2/xyz, and fr3/office, with approximately 500 image frames from each scene, and the sampling strides are 1, 1, 2, 6, and 5, respectively. For the Replica dataset, we sampled 500 frames for all scenes with a stride of 4.

\begin{figure}[!h]
    \centering
    \includegraphics[width=0.7\linewidth]{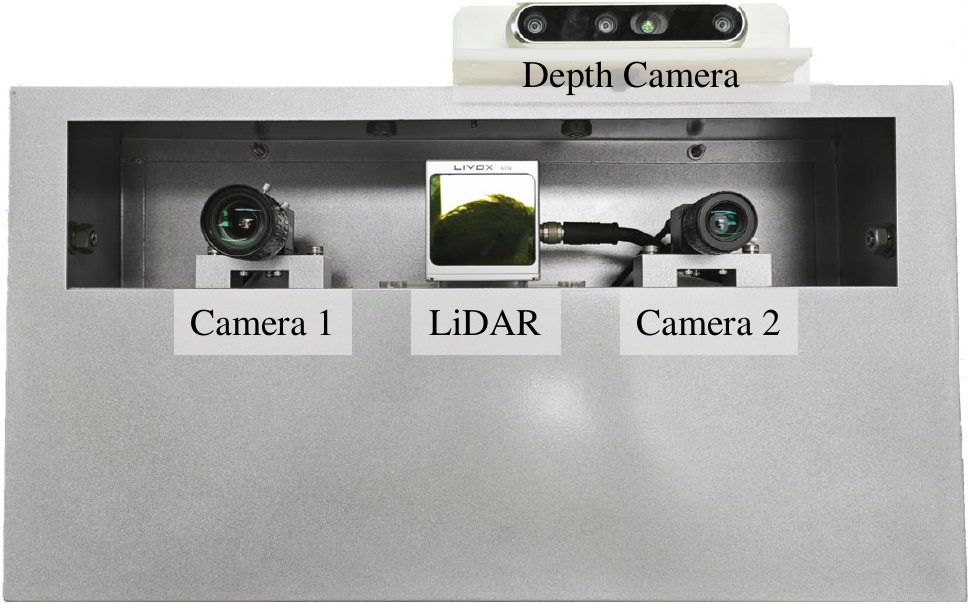}
    \caption{Sensor pod with synchronized LiDAR, RGB, and depth cameras.}
    \label{fig:sensor_pod}
\end{figure}

\begin{figure}
    \centering
    \includegraphics[width=1\linewidth]{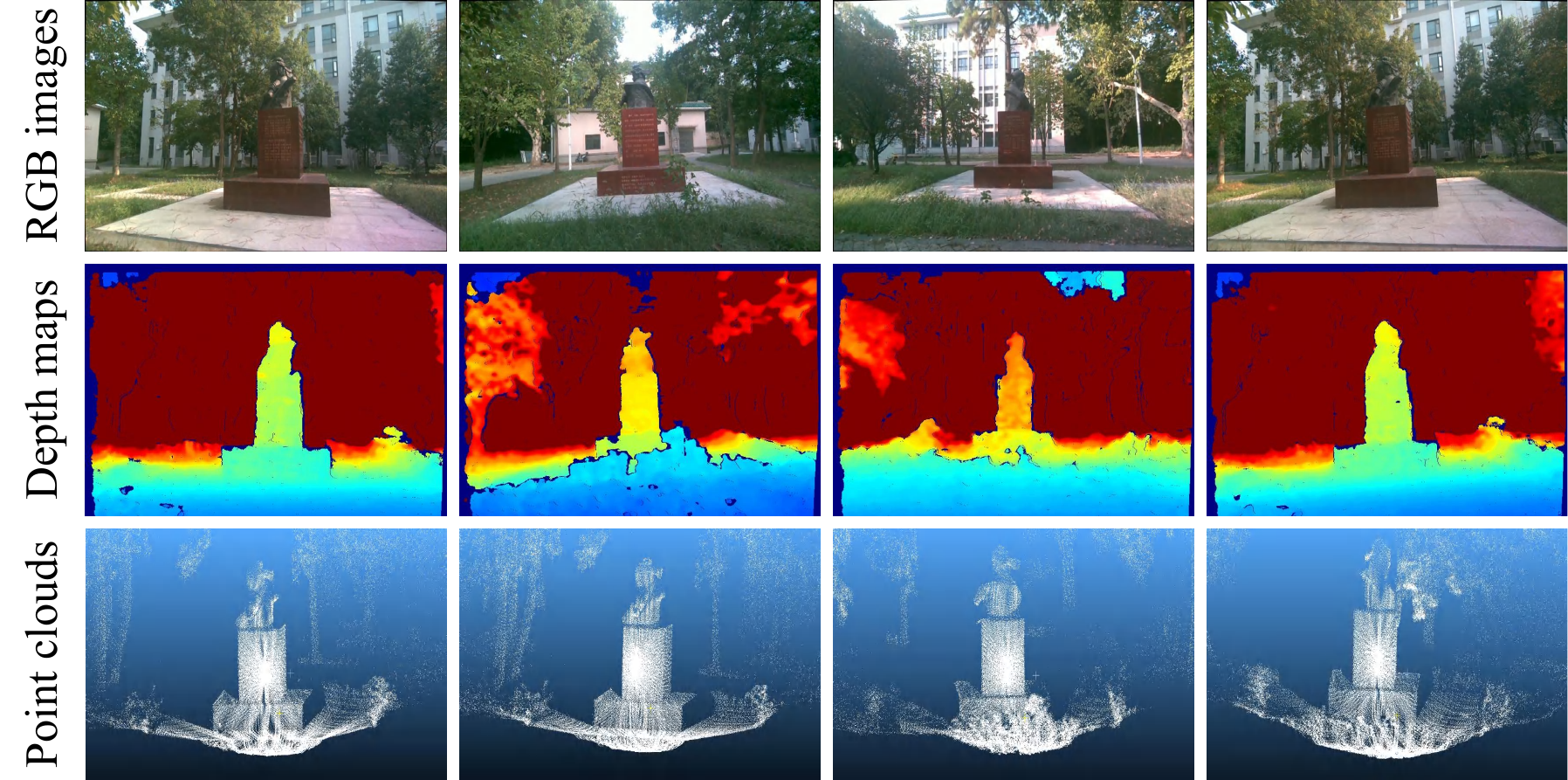}
    \caption{Self-collected data examples.}
    \label{fig:recored_data_example}
\end{figure}

To further demonstrate the generalization ability of the proposed LiGSM, we build our own sensor pod to capture real-world data, as shown in Fig. \ref{fig:sensor_pod}. The sensor pod is equipped with two RGB cameras, a depth camera, and a LiDAR. The camera is \textit{Huarui A7500CU35 Industrial Camera}, the RGB-D camera is \textit{Intel RealSense Depth Camera D455}, and the LiDAR is \textit{Livox Avia}. These sensors are all synchronized and calibrated as prerequisites. 
We collect a typical scenario with a total of 90 stations to evaluate the pose estimation and image rendering performance. Fig. \ref{fig:recored_data_example} presents data examples, where the first and second rows are RGB images and the corresponding depth images captured by the Realsense depth camera, respectively. The third row shows the paired LiDAR point clouds. Our method uses RGB images and LiDAR point clouds as input, while SplaTAM \cite{keetha2024splatam} uses RGB images and depth images as input. 

\subsection{Experimental Setup and Evaluation Metrics}
The runtime environment for LiGSM consists of an Intel(R) Xeon(R) Platinum 8336C CPU @ 2.30GHz, coupled with an NVIDIA GeForce RTX 4090 GPU. The parameter settings are as follows: $r_s, \lambda_c, \lambda_l$ and $\lambda_j$ are set to 0.3, 1.0, 20.0, and 10.0, respectively.
$\lambda_{color}, \lambda_{depth}, \lambda_{ssim}$ follow the SplaTAM \cite{keetha2024splatam} as 0.5, 1.0, and 0.2, respectively. 

For effectiveness demonstration, we select SplaTAM \cite{keetha2024splatam}, Point-SLAM \cite{sandstrom2023point}, and NICE-SLAM \cite{zhu2022nice} as competitors. To evaluate these methods, we use the absolute trajectory error (ATE) \cite{sturm2012tum} to assess pose tracking accuracy, and PSNR, SSIM \cite{wang2004image}, and LPIPS \cite{zhang2018unreasonable} to evaluate image rendering quality. PSNR quantifies image quality via the signal-to-noise ratio, highlighting the signal's maximum power relative to noise. SSIM better captures visual similarity by accounting for luminance, contrast, and structure. LPIPS further advances this by assessing similarity through learned perceptual features, detecting subtleties overlooked by PSNR and SSIM. The rendering metrics are calculated based on the full-resolution images rendered under the estimated trajectory, similar to Point-SLAM \cite{sandstrom2023point}.

\subsection{Tracking Performance}
Tab. \ref{tab: exp_tum} presents the pose tracking results on the TUM RGB-D dataset \cite{sturm2012tum}. LiGSM achieves an average ATE \cite{sturm2012tum} of 4.77 cm, significantly outperforming other methods. This is nearly half of the error of SplaTAM \cite{keetha2024splatam} with 8.36 cm, highlighting the effectiveness of LiGSM on pose estimation.

\begin{table}[h]\scriptsize
    \centering
    \setlength{\tabcolsep}{2.5pt}
    \caption{Tracking ATE on TUM RGB-D dataset [cm]}
    \label{tab: exp_tum}
    \begin{tabular}[t]{l|c|c|c|c|c|c}
        \toprule
        Method &  Avg. & fr1/desk & fr1/desk2 & fr1/room & fr2/xyz & fr3/office \\
        \midrule
        NICE-SLAM \cite{zhu2022nice} & 24.91 & 3.41 & 4.84 & 104.77 & 2.28 & 9.27 \\
        \midrule
        Point-SLAM \cite{sandstrom2023point} & 7.65 & 2.73 & 4.64 & 23.48 & 1.65 & 5.73  \\
        \midrule
        SplaTAM \cite{keetha2024splatam} & 8.36 & 3.35 & 6.55 & 14.16 & 1.35 & 16.40  \\
        \midrule
        LiGSM & \textbf{4.77} & \textbf{2.31} & \textbf{3.36} & \textbf{13.11} & \textbf{1.08} & \textbf{3.98}  \\
        \bottomrule
    \end{tabular}\vspace{-0mm}
\end{table}

\subsection{Rendering Performance}
\begin{figure}
    \centering
    \includegraphics[width=1\linewidth]{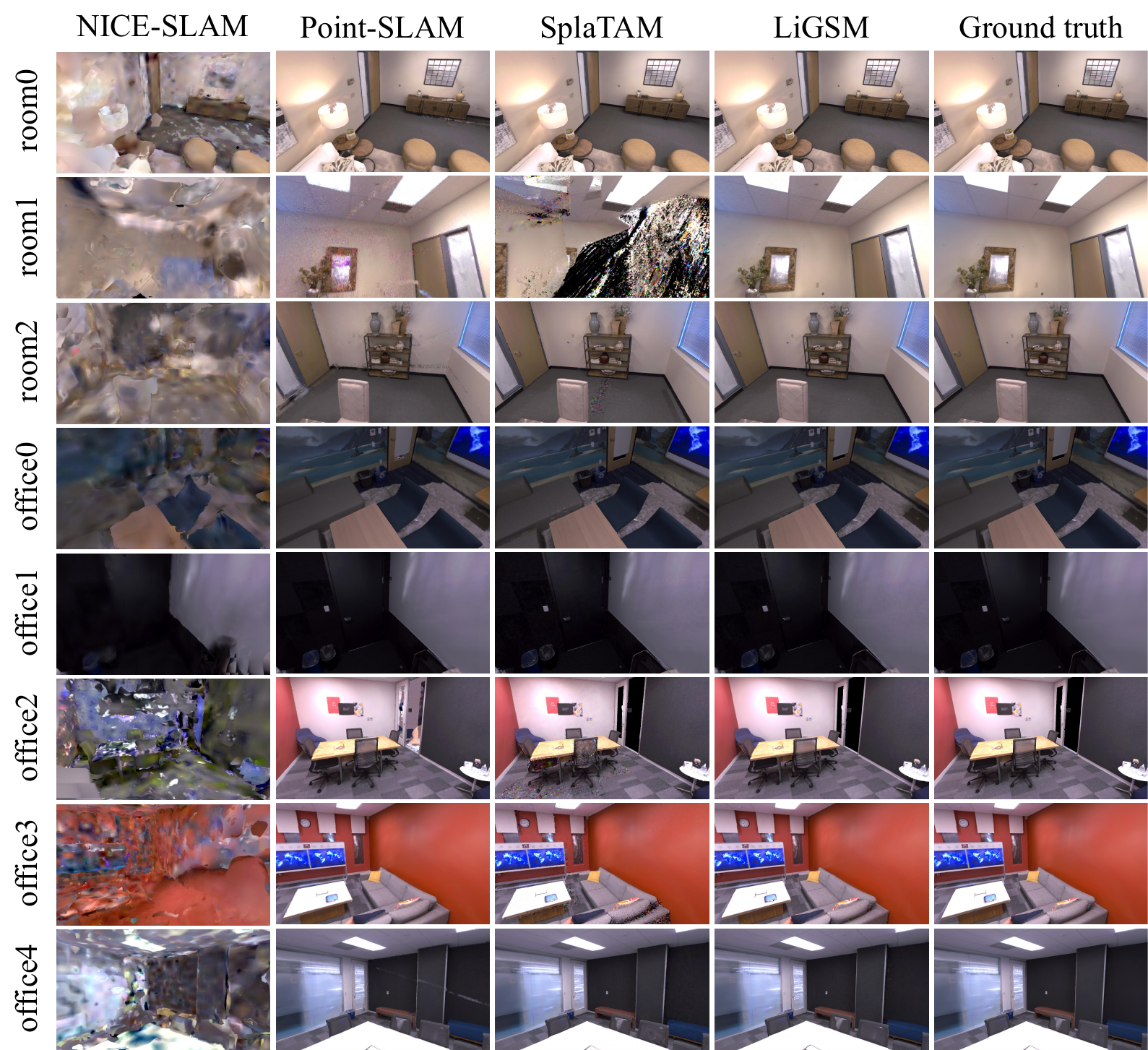}
    \caption{Rendering performance on Replica.}
    \label{fig:exp_replica}
\end{figure}

Tab. \ref{tab: exp_replica} displays the rendering results on the synthetic Replica dataset \cite{straub2019replica}.
It can be observed that LiGSM obtains the best performance in image rendering quality, as evidenced by its highest PSNR values for pixel-wise accuracy, top SSIM scores for structural similarity, and the lowest LPIPS values for perceptual closeness to the ground truth. In comparison to other scenes, the results of other methods have deteriorated in the room1 data, but LiGSM still maintains high-quality rendering results, indicating that our method is robust across different scenes. Fig. \ref{fig:exp_replica} shows several samples of the rendering results. LiGSM shows superior rendering quality across all scenes. In contrast, SplaTAM exhibits subpar results in the room1 scenario. Point-slam struggles with capturing high-frequency details in the rendered images, and NICE-SLAM produces inferior renderings. Both quantitative and qualitative results demonstrate that LiGSM outperforms competitors in image rendering.

\begin{table*}[h]
    \centering
    \vspace{3mm}
    \caption{Rendering performance on Replica}
    \label{tab: exp_replica}
    \begin{tabular}[t]{l|c|c|c|c|c|c|c|c|c|c}
        \toprule
        Method & Metric &  Avg. & room0 & room1 & room2 & office0 & office1 & office2 & office3 & office4 \\
        \midrule
        \multirow{3}{*}{NICE-SLAM \cite{zhu2022nice}} & PSNR $\uparrow$ & 15.47 & 13.8 & 14.23 & 15.18 & 19.75 & 26.4 & 9.35 & 12.24 & 12.82 \\
        & SSIM $\uparrow$ & 0.522 & 0.41 & 0.456 & 0.482 & 0.607 & 0.859 & 0.471 & 0.363 & 0.526 \\
        & LPIPS $\downarrow$ & 0.596 & 0.671 & 0.661 & 0.627 & 0.534 & 0.34 & 0.67 & 0.668 & 0.598 \\
        \midrule
        \multirow{3}{*}{Point-SLAM \cite{sandstrom2023point}} & PSNR $\uparrow$ & 30.36 & 24.5 & 24.71 & 28.98 & 35.23 & 37.74 & 28.95 & 31.68 & 31.1 \\
        & SSIM $\uparrow$ & 0.92 & 0.859 & 0.802 & 0.928 & 0.964 & 0.982 & 0.925 & 0.95 & 0.952 \\
        & LPIPS $\downarrow$ & 0.221 & 0.275 & 0.417 & 0.212 & 0.146 & 0.133 & 0.217 & 0.161 & 0.208 \\
        \midrule
        \multirow{3}{*}{SplaTAM \cite{keetha2024splatam}} & PSNR $\uparrow$ & 32.06 & 31.22 & 18.42 & 31.76 & 38.38 & \textbf{39.3} & 31.5 & 32.19 & 33.74 \\
        & SSIM $\uparrow$ & 0.918 & 0.945 & 0.633 & 0.934 & 0.976 & 0.982 & 0.952 & 0.967 & 0.959 \\
        & LPIPS $\downarrow$ & 0.139 & 0.085 & 0.454 & 0.111 & 0.072 & 0.07 & 0.105 & 0.089 & 0.13 \\
        \midrule
        \multirow{3}{*}{LiGSM (ours)} & PSNR $\uparrow$ & \textbf{34.48} & \textbf{31.89} & \textbf{33.26} & \textbf{32.40} & \textbf{40.28} & 38.08 & \textbf{32.03} & \textbf{33.58} & \textbf{34.34} \\
        & SSIM $\uparrow$ & \textbf{0.98} & \textbf{0.972} & \textbf{0.972} & \textbf{0.972} & \textbf{0.99} & \textbf{0.984} & \textbf{0.98} & \textbf{0.986} & \textbf{0.982} \\
        & LPIPS $\downarrow$ & \textbf{0.068} & \textbf{0.067} & \textbf{0.068} & \textbf{0.08} & \textbf{0.044} & \textbf{0.069} & \textbf{0.074} & \textbf{0.059} & \textbf{0.086} \\
        \bottomrule
    \end{tabular}\vspace{-0mm}
\end{table*}

\subsection{Performance on Self-Collected Dataset}
Tab. \ref{tab: exp_recored} shows the evaluation results on the self-collected data. LiGSM uses the original extrinsic parameters from LiDAR-camera calibration \cite{yuan2021pixel}, LiGSM* uses optimized extrinsics from our pipeline, SplaTAM runs the original algorithm, and SplaTAM* \cite{keetha2024splatam} uses the poses from LiGSM. 
It can be observed that the proposed LiGSM achieves the best in all metrics when using the optimized extrinsic parameters. Comparing SplaTAM* and SplaTAM, SplaTAM* outperforms SplaTAM in all evaluation metrics, indicating that the poses obtained by the LiGSM are more accurate than those of SplaTAM.

\begin{table}[h]
    \centering
    \caption{Performance on the self-collected dataset}
    \label{tab: exp_recored}
    \begin{tabular}[t]{l|c|c|c|c}
        \toprule
        Metric & SplaTAM & SplaTAM* & LiGSM & LiGSM* \\
        \midrule
        PSNR $\uparrow$ & 15.71 & 19.16 & \textbf{20.09} & 19.03 \\
        SSIM $\uparrow$ & 0.572 & 0.760 & \textbf{0.815} & 0.785 \\
        LPIPS $\downarrow$ & 0.363 & 0.251 & \textbf{0.171} & 0.197 \\
        \bottomrule
    \end{tabular}\vspace{-2mm}
\end{table}

\begin{figure}[h]
    \centering
    \includegraphics[width=1\linewidth]{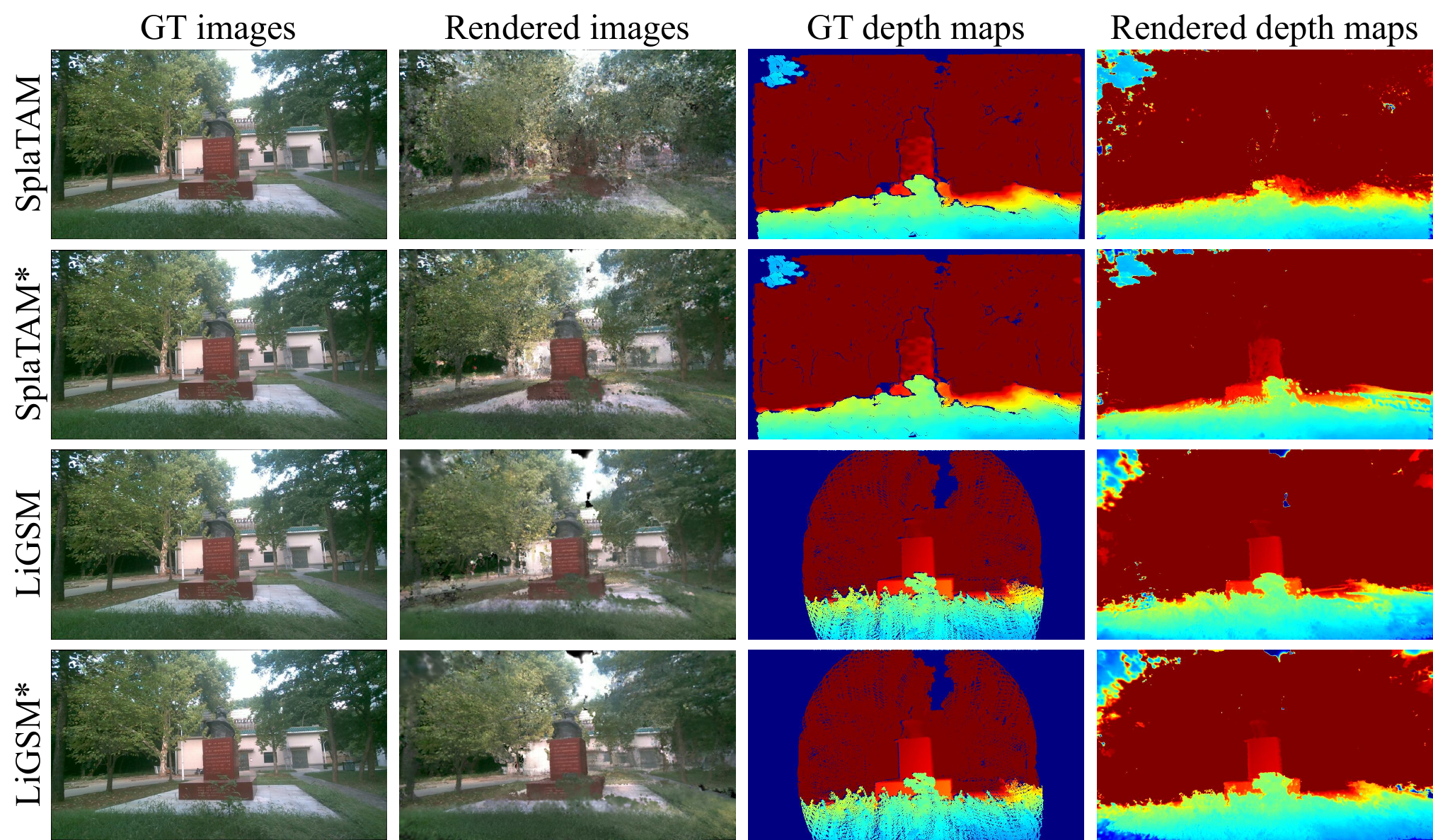}
    \caption{RGB and depth rendering on the self-collected data.}
    \label{fig:plot-recored}
\end{figure}

Fig. \ref{fig:plot-recored} displays the rendering results. The ground truth depth maps in SplaTAM and SplaTAM* come from a depth camera, and those in LiGSM and LiGSM* come from the point cloud projection. The ground truth depth map derived from the projection of LiDAR point clouds is sparser than that from the RGB-D camera, with some areas in the image lacking depth information. It can be observed that the SplaTAM* with optimized poses yields better results compared to the original SplaTAM.  
 Although using a sparse depth input, LiGSM still shows the best depth rendering with dense depth information and the highest similarity to the ground truth, indicating the effectiveness of LiGSM in using sparse LiDAR points. Furthermore, LiGSM completes the missed depth information without LiDAR point input, which further demonstrates its robustness. 


\subsection{Extrinsics Estimation Performance}

Table \ref{tab: exp_recored} demonstrates that LiGSM*, utilizing optimized extrinsic parameter, exhibits performance that is subsequent only to LiGSM employing the original calibrated extrinsic. Compared to SplaTAM*, LiGSM* has a similar rendering effect, slightly lower PSNR, but higher SSIM and LPIPS.

\begin{figure}[h]
    \centering
    \includegraphics[width=0.9\linewidth]{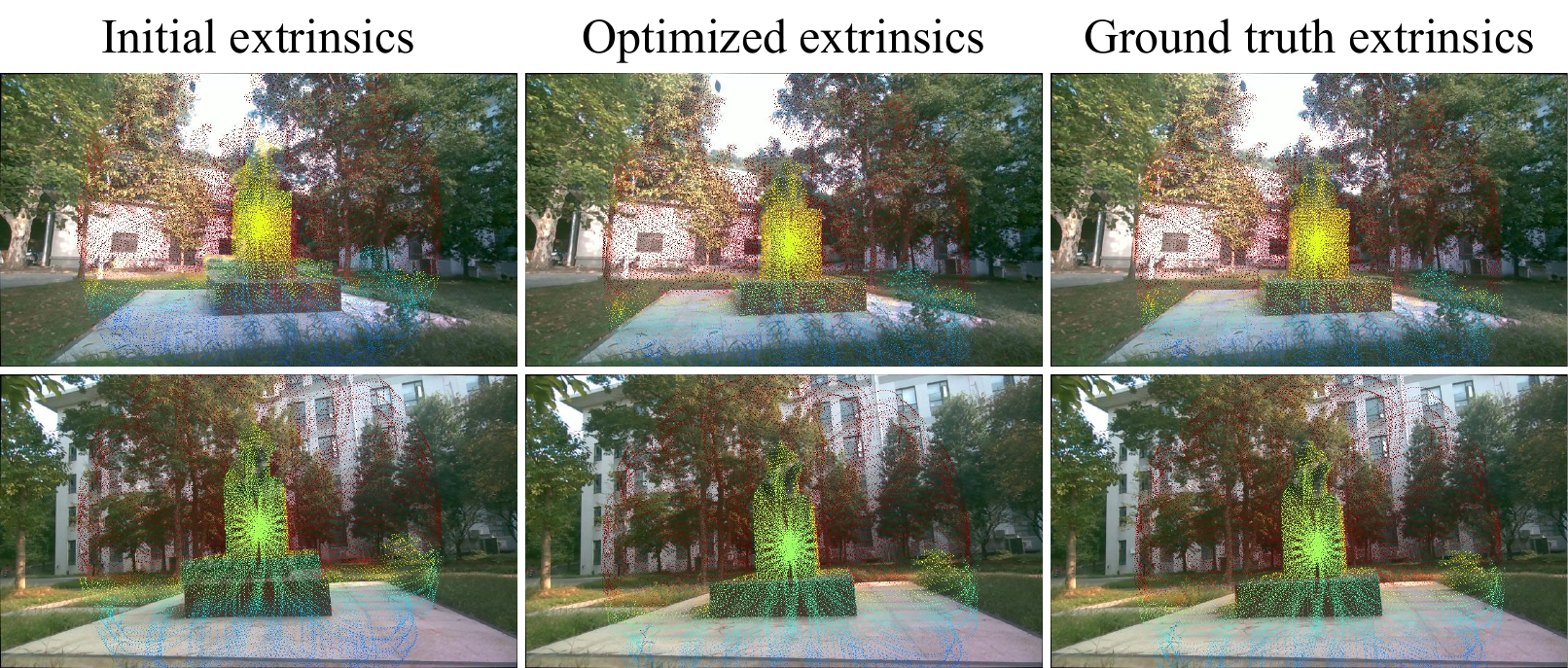}
    \caption{RGB image overlays with the LiDAR point cloud projections using initial extrinsic, optimized and ground truth extrinsic parameters, respectively.}
    \label{fig:extrinsic_opt}
\end{figure}

To visualize the effects of extrinsic optimization, we use the extrinsic parameter to transform the LiDAR point cloud into the camera coordinate system and project it in the image plane to overlay on the RGB image. Fig. \ref{fig:extrinsic_opt} illustrates the overlayed images using an initial inaccurate extrinsic parameter (left), the optimized extrinsics (mid), and the ground truth extrinsics (right). It can be observed that the projected points are initially with noticeable misalignment with the RGB images, but after extrinsic optimization, they become essentially aligned with the points in the RGB image, and the alignment is consistent with the ground truth.

\section{Conclusions and Future Work}
\label{sec:conclusion}

In this paper, we propose a LiDAR-enhanced 3DGS-based mapping pipeline that integrates the geometric information of LiDAR point clouds with the textural information of images to achieve high-precision and high-quality 3D scene mapping. By optimizing the joint observation of images and point clouds, we achieve robust LiDAR-camera extrinsic parameter optimization and pose estimation. Experiments on public and self-collected datasets demonstrate the effectiveness of our method in tracking, mapping, and extrinsic optimization. In the future, we will explore the scalability of LiGSM in large-scale scenarios, enhancing its ability for efficient and meticulous mapping. Besides, we are committed to improving the computation speed of LiGSM, ensuring that it can perform efficient 3D mapping to meet the demands of real-time applications.

\section{Acknowledgement}

This work was supported by the NSFC grants under contracts Nos. 62301370 and 62325111, the NSF of Hubei Province, China under No.JCZRYB202500481, and Wuhan Natural Science Foundation Exploration Program under No.2024040801020233.

\bibliographystyle{IEEEtran.bst}
\bibliography{references}

\end{document}